\documentclass{article}

\usepackage{arxiv}

\usepackage[utf8]{inputenc} 
\usepackage[T1]{fontenc}    
\usepackage{hyperref}       
\usepackage{url}            
\usepackage{booktabs}       
\usepackage{amsfonts}       
\usepackage{nicefrac}       
\usepackage{microtype}      

\usepackage{graphicx}
\newcommand{\ourmethod}{$PICO$\xspace}
\usepackage{amssymb}
\usepackage{amsmath}
\usepackage{xspace}
\usepackage{xcolor}
\usepackage{subfig}
\usepackage{comment}
\usepackage{cite}
\usepackage{url}

\newcommand{\argmaxA}{\mathop{\mathrm{argmax}}\limits}   

\title{\LARGE \bf
PICO: Primitive Imitation for COntrol 
}

\author{Corban G. Rivera, Katie M. Popek, Chace Ashcraft, Edward W. Staley, \\ \textbf{  Kapil D. Katyal, and Bart L. Paulhamus } \\  
Intelligent Systems Center\\ Johns Hopkins University Applied Physics Lab\\
        11100 Johns Hopkins Rd. Laurel, MD 20723 \\
        {\tt\small corban.rivera@jhuapl.edu}%
}

\begin{document}

\maketitle
\thispagestyle{empty}
\pagestyle{empty}

\begin{abstract}

In this work, we explore a novel framework for control of complex systems called Primitive Imitation for Control (\ourmethod). The approach combines  ideas from imitation learning, task decomposition, and novel task sequencing to generalize from demonstrations to new behaviors.  Demonstrations are automatically decomposed into existing or missing sub-behaviors which allows the framework to identify novel behaviors while not duplicating existing behaviors.  Generalization to new tasks is achieved through dynamic blending of behavior primitives.  We evaluated the approach using demonstrations from two different robotic platforms.  The experimental results show that \ourmethod is able to detect the presence of a novel behavior primitive and build the missing control policy.

\end{abstract}

\section{INTRODUCTION}

Human-robotic fusions controlled through brain computer interfaces (BCI) have tremendous potential to impact human health and capabilities through applications like intelligent motor and cognitive prostheses~\cite{guger1999prosthetic, muller2007control, mcfarland2008brain, fifer2013simultaneous, hotson2016individual, Zhao2017, Akinola2017}. BCI-based control approaches are currently limited by the number of independent degrees of freedom that can be reliability controlled directly.  The number of degrees of freedom on motor prostheses for example can be several dozen~\cite{pasquina2015recent}.  This is about an order of magnitude more degrees of freedom than can be reliability generated by current non-invasive BCI systems.  The dilemma is how to control high-degree of freedom complex machines with only a few control inputs. 

One approach to increase the impact of limited control inputs is through modularity and hierarchical control mechanisms~\cite{Zhao2017, Akinola2017}.  The idea is to use the limited number of inputs to select a primitive control policy, from a library of primitive behaviors, and potentially a target. Complex tasks are performed by chaining primitive behaviors.

\begin{figure}[ht!]
\centering
  \includegraphics[width=.8\linewidth]{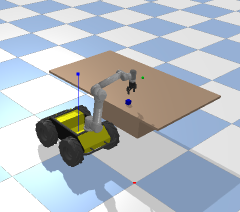}
  \caption{Husky-UR5 Reach and Grasp Environment }
  \label{fig:husky}
\end{figure}

As an example of this scenario, consider a Universal Robots UR5~\cite{UR} manipulator mounted on a Clearpath Husky platform~\cite{Husky} as shown in Fig.~\ref{fig:husky}.  The UR5 is used to demonstrate reaching, grabbing, and lifting a block on a table.  Other tasks may require performing these actions in another order, so it may be useful to learn and maintain a collection of these primitive behaviors for later use. While the underlying behavior primitives are well defined for the reach-and-grasp scenario, other example scenarios may not have as well defined or labeled primitives.  In this work, we assume that the underlying label of the behaviors shown in the task demonstrations is unknown.  

The questions we investigate are how might we learn and maintain the primitive library from unlabeled demonstrations and, assuming the behavior primitive library exists, how would one know when to use, adapt, or create a new primitive behavior.  We propose that the behavior library should be actively maintained to minimize redundancy and maximize the ability to reconstruct complex tasks through chains of the primitive behaviors.  In this work, we explore techniques to directly optimize for these criteria by building on methods that learn from demonstration.

We explore maintaining a behavior primitive library in an online learning scenario.  Given a potentially non-empty behavior primitive library and a new set of unlabeled task demonstrations, we seek to update the behavior primitive library to maximally accommodate the new demonstrations while maintaining the ability to reconstruct previously demonstrated trajectories.

\begin{figure}[ht]
\centering
  \includegraphics[width=1\linewidth]{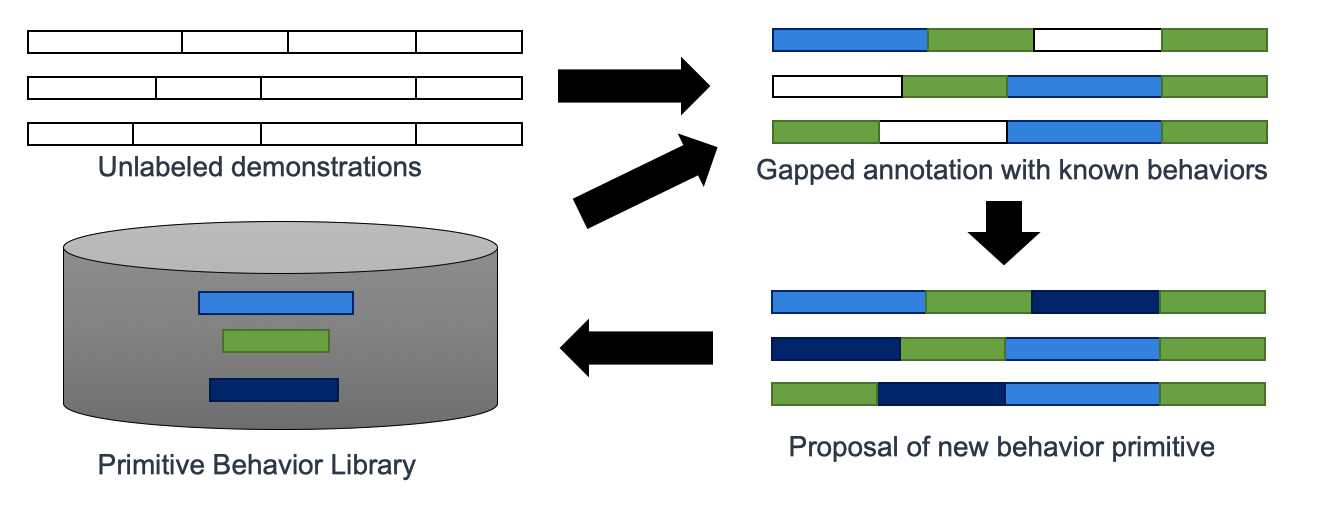}
  \caption{An overview of \ourmethod. The approach takes as input unlabeled demonstrations and a library of primitive behaviors.  The goal is to predict the primitive behavior label associated with each time point in all demonstrations. Additional behavior primitive models can be trained to fill gaps that are not well represented by existing behavior primitives. }
  \label{fig:overview}
\end{figure}

Our contribution is an approach called \ourmethod that simultaneously learns subtask decomposition from unlabeled task demonstrations, trains behavior primitives, and learns a hierarchical control mechanism that allows blending of primitive behaviors to create even greater behavioral diversity, an overview is shown in Fig.~\ref{fig:overview}. Our approach directly optimizes the contents of the primitive library to maximize the ability to reconstruct unlabeled task demonstrations from sequences of primitive behaviors.

\section{PRELIMINARIES}


Learning from demonstration (LfD) and imitation learning allow agents to execute a task by observing the task being performed \cite{Hussein:2017:ILS:3071073.3054912}. In the robotics domain, a goal of imitation learning is to produce a mapping, $\pi$, from states to actions, known as a control \emph{policy} \cite{ARGALL2009469, schaal2010learning}, that has the maximum likelihood of producing the demonstration dataset $\mathcal{D} = \{\rho_1,\rho_2,\dots,\rho_n\}$, where each $\rho = ((s_1,a_1),(s_2,a_2),\dots,(s_T,a_T)$ is a demonstration trajectory of of state, action pairs. The demonstrations can be created by another control policy \cite{distillation}, by a human expert \cite{konidaris2012}, or in a simulated environment \cite{TACO18, compile2019}. Let $\pi_\theta$ parameterized by $\theta$. The goal is then to optimize Equation \ref{bc} by varying $\theta$.

\begin{equation}
    \max \mathbb{E}_\rho[\sum_{t=1}^T\log\pi_\theta(a_t|s_t)]\label{bc}
\end{equation}

Following optimization, covariate drift can cause errors in the control process that can place the robot in a previously unobserved state. Control policies will have higher action prediction errors in parts of the state space that it has not observed, leading to poor action predictions and compounding errors with increased iterations of the policy.  One approach that has been introduced to decrease the impact of covariate shift is to introduce noise into the demonstrations used for learning \cite{DART}.  This approach increases the amount of state space covered by the policy and improves action predictions around the demonstrations, leading to better generalization and error tolerance.  

\subsection{Model Agnostic Meta-Learning}
In meta-learning a model is trained on a variety of learning tasks and the parameters of the method are fine-tuned for generalization. The idea of meta-learning is to combine a set of learner models to improve performance on a task more quickly than one without pretrained models.  This is a common strategy for one-shot~\cite{santoro2016oneshot} or few shot scenarios, where a model must be trained using one or a few examples. Some approaches for meta-learning come from the reinforcement learning~\cite{Finn2017}, which typically differ in how they update individual learners.  Some meta-learning methods update models using gradient information~\cite{Finn2017} and others learn how to update learners from data~\cite{learning_to_learn,Bengio2002}.

\section{RELATED WORK}

Imitation learning alone does not provide a mechanism to generalize demonstrations to new tasks.  One mechanism to address this challenge is task decomposition, which has the goal of identifying subtasks from demonstration.  Subtasks can be made into sub-policies through imitation learning, including methods methods that combine subtask discovery with imitation learning~\cite{TACO18,NTP18}. By decomposing demonstrations into subtasks, it becomes possible to permute the sequence of sub-policies to achieve greater task diversity and generalizability. However, decomposing demonstrations into subtasks that are maximally useful for recombination is a challenge in task decomposition~\cite{TACO18}.

Once sub-task policies are established, a hierarchical control policy can be learned that identifies the sequence of policies needed to achieve a specified goal. Given a sufficiently diverse set of demonstrations the reasoning layer can be learned from a set of demonstrations~\cite{NTP18}. Several approaches for learning hierarchical architectures for control policies from limited demonstrations have been proposed~\cite{TACO18, NTP18, duan2017}.  We were inspired by the work on mixtures-of-experts\cite{mixture_of_experts,experts}
which includes a similar hierarchical representation.

Some approaches assume that the behavior primitive library is fully trained in advance~\cite{NTP18}.  In the reinforcement learning domain, the options framework~\cite{drl_options, andreas2017, kulkarni2016} and hierarchical reinforcement learning~\cite{HRL} are common approaches for organising hierarchies of policies.  The techniques in reinforcement learning are often predicated on being able to interact with an environment and collect a lot of data.  In this work, we focus on learning hierarchical task decomposition strategies from a limited set of demonstrations.

\subsection{Task Sketch for Sub-policy Discovery}
Some related approaches~\cite{andreas2017, mu2019plots} perform demonstration decomposition by combining both demonstrations and task sketches.  The literature refers to these approaches as \textit{weakly-supervised} because the order of tasks is given and the exact transition points within a demonstration must be inferred. 

Let $\mathcal{D}$ be our dataset containing
trajectories $\rho = ((s_0,a_0),(s_1,a_1),\ldots,((s_T,a_T))$ of length $T$ containing state-action tuples $(s,a)$ for state $s$ and action $a$.
Given a library of sub-tasks policies $\mathcal{B}=(\pi_1,\pi_2,\ldots,\pi_K)$,  A task sketch $\tau =(\tau_1,\tau_2,\ldots,\tau_L$) is a sequence of sub-tasks labels where $L$ is the length of the sketch.  A path is a sequence of sub-task labels $\zeta = (\zeta_1,\zeta_2,\ldots,\zeta_T)$ where $T$ is the length of a demonstration.  We assume that $L<<T$.  We say that a path $\zeta$ matches a task sketch $\tau$ if $\tau = \zeta $ after  removing all adjacent duplicate sub-tasks in $\zeta$.  For example, the path $(\pi_2,\pi_2,\pi_2,\pi_3,\pi_3,\pi_1,\pi_1,\pi_1,\pi_1)$ matches the task sketch $(\pi_2,\pi_3,\pi_1)$.

\section{METHODS}

In this section we describe the approaches most closely aligned with our work referred to as CTC~\cite{CTC06} and TACO~\cite{TACO18}.  Then, we introduce our approach called Primitive Imitation for COntrol (\ourmethod).

\subsection{Connectionist Temporal Classification}
Given a dataset $\mathcal{D}$ and task sketch $\tau$, one approach to obtain a set of generalizable sub-tasks $\mathcal{B}$ is to separately learn alignment of trajectories to the task sketch then learn the control policies for sub-tasks with behavior cloning.
Connectionist Temporal Classification (CTC)~\cite{CTC06} addresses the problem of aligning sequences of dissimilar lengths. There are potentially multiple ways in which a path could be aligned to a task sketch. Let $\mathbb{Z}_{(T,\tau)}$ be the set of all paths of length $T$ that match the task sketch $\tau$.  The CTC objective maximises the
probability of the task sketch $\tau$ given the input trajectory $\rho$:

\begin{equation}
   \theta^* =  \argmaxA_\theta \mathbb{E}_{(\rho,\tau)}[p_\theta(\tau | \rho)]
\label{eqn:ctc1} 
\end{equation}

\begin{equation}
    \theta^* =  \argmaxA_\theta \mathbb{E}_{(\rho,\tau)}[\sum_{\zeta \in \mathbb{Z}_{(T,\tau)}} \prod_{t=1}^T p_\theta(\zeta_t,|\rho_t)]
    \label{eqn:ctc2}
\end{equation}

$p_\theta(\zeta_t,|\rho)$ is commonly represented as a neural network with parameters $\theta$ that outputs the probability of each sub-task policy in $\mathcal{B}$.  The objective is solved efficiently using dynamic programming. Inference using the neural network model is used to find a maximum likelihood path $\zeta$ for a trajectory $\rho$. The labels in $\zeta$ provide an association between state-action tuples $(s_t,a_t)$ and subtask policies $\pi \in \mathcal{B}$.  The state-action policies associated with a single sub-task are used to create a sub-task policy using behavior cloning.

\subsection{Temporal Alignment for Control} Given a demonstration $\rho$ and a task sketch $\tau$, Temporal Alignment for Control (TACO)~\cite{TACO18} will learn where each subtask begins and ends in the trajectory and simultaneously trains a library of sub-tasks policies $\mathcal{B}$.  TACO maximizes the joint log likelihood of the task sequence $\tau$ and the actions from sub-task policies contained in $\mathcal{B}$ conditioned on the states. Let $\textbf{a}_\rho$ and $\textbf{s}_\rho$ be the set of actions and states respectively in trajectory $\rho$.

\begin{equation}
    p(\tau,\textbf{a}_\rho|\textbf{s}_\rho) = \sum_{\zeta \in \mathbb{Z}_{(T,\tau)}} p(\zeta|\textbf{s}_\rho) \prod_{t=1}^T \pi_{\zeta_t}(a_t|s_t)
    \label{eqn:taco}
\end{equation}

where $p(\zeta|\textbf{s}_\rho) $ is the product of action probabilities associated with any given path $\zeta$. The path $\zeta$ determines which data within $\rho$ corresponds to each sub-task policy $\pi$ and  $\prod_{t=1}^T \pi_{\zeta_t}(a_t|s_t)$ is the behavior cloning objective from Equation \ref{bc}.

\subsection{Primitive Imitation for Control (PICO)}
In this work, we introduce Primitive Imitation for Control. (\ourmethod). The approach differs from previous work in a few important ways. \ourmethod  similarly decomposes behavior primitives from demonstration. It optimizes the action conditioned on state and does not require a task sketch, and unlike CTC\cite{CTC06}, our approach simultaneously learns to segment demonstrations and trains underlying behavior primitive models. 

We aim to reconstruct the given trajectories as well as possible using the existing sub-task policy library.   As shown in Equation \ref{mse}, we seek to minimize the sum of squared error between the observed action and the predicted action for all actions over all timepoints $T$ and all trajectories $\rho \in \mathcal{D}$.   We refer to this objective as minimizing reconstruction error. Let  $(s_{\rho_t},a_{\rho_t})$ be the state-action tuple corresponding to $\rho_t$ timepoint $t$ in trajectory $\rho$.  The action prediction, equation \ref{action_prediction}, is the product of the probability $p(\pi|s_{\rho_t})$ of a sub-task policy $\pi$ conditioned on the state $s_{\rho_t}$ and the action predicted by policy $\pi(s_{\rho_t})$ for the state $s_{\rho_t}$.  Substituting equation \ref{action_prediction} into \ref{mse} results in Equation \ref{mse2} which is the optimization problem for \ourmethod.

\begin{equation}
\min \sum_{\rho \in \mathcal{D}} \sum_{t=0}^T (a_{\rho_t}-\hat{a}_{\rho_t})^2
\label{mse}
\end{equation}

\begin{equation}
    \hat{a}_{\rho_t} = \sum_{\pi \in \mathcal{B}} p(\pi|s_{\rho_t}) \pi(s_{\rho_t} )
    \label{action_prediction}
\end{equation}

\begin{equation}
\min \sum_{\rho \in \mathcal{D}} \sum_{t=0}^T (a_{\rho_t}-
\sum_{\pi \in \mathcal{B}} p(\pi|s_{\rho_t}) \pi(s_{\rho_t} 
))^2
\label{mse2}
\end{equation}

\subsection{Neural Network Architecture}

\begin{figure}[ht]
\centering
  \includegraphics[width=1\linewidth]{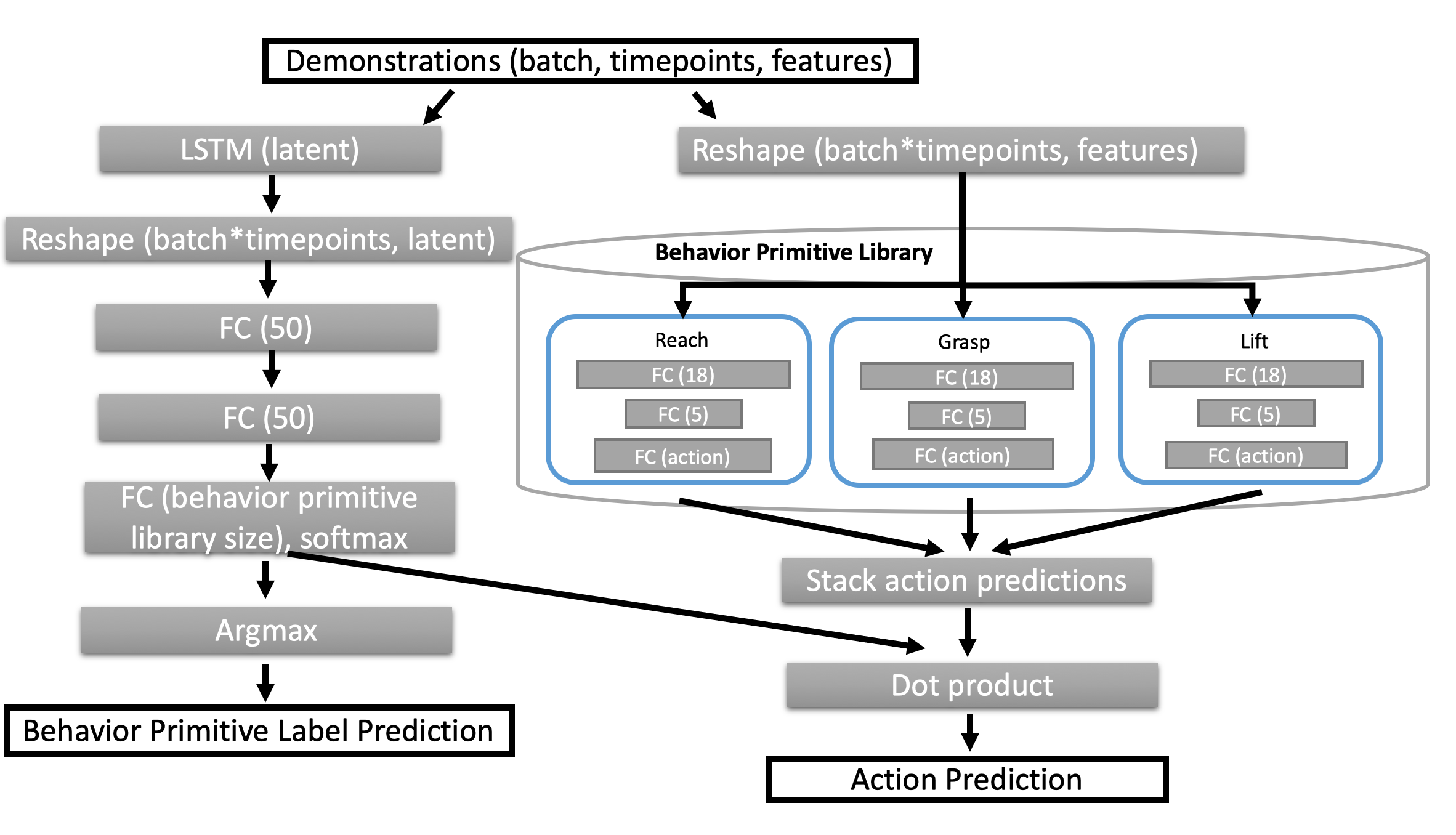}
  \caption{Neural network architecture for \ourmethod.  Given a set of input trajectories and a behavior primitive library, the core architecture follows two branches, the left most branch estimates a distribution over the behavior primitives.  The right hand branch estimates the action prediction from each primitive behavior sub-model.  We compute the predicted action as a linear combination between the behavior primitive distribution and the set of predicted actions from all behavior primitives.  }
  \label{fig:model}
\end{figure}

Estimates of both $p(\pi|s_{\rho_t})$ and $\pi(s_{\rho_t})$ are given by a recurrent neural network architecture. Figure \ref{fig:model} gives an overview of the recurrent and hierarchical network architecture. 
We solve for the objective in Equation \ref{mse2} directly by back propagation through a recurrent neural network with equation \ref{mse} as the loss function.  The model architecture is composed of two branches that are recombined to compute the action prediction at each timepoint.  

To more easily compare with other approaches that do not blend sub-task policies, we estimate the maximum likelihood sub-task policy label at each timepoint. We refer to sub-task policies as behavior primitives.  The behavior primitive label prediction is given by the maximum likelihood estimate of $\pi$ shown in Equation \ref{eqn:label} for time $t$ in trajectory $\rho$. 

\begin{equation}
     \argmaxA_{\pi \in \mathcal{B}} p(\pi|\rho_t)
    \label{eqn:label}
\end{equation}

\begin{figure}[ht]
\centering
  \includegraphics[width=1\linewidth]{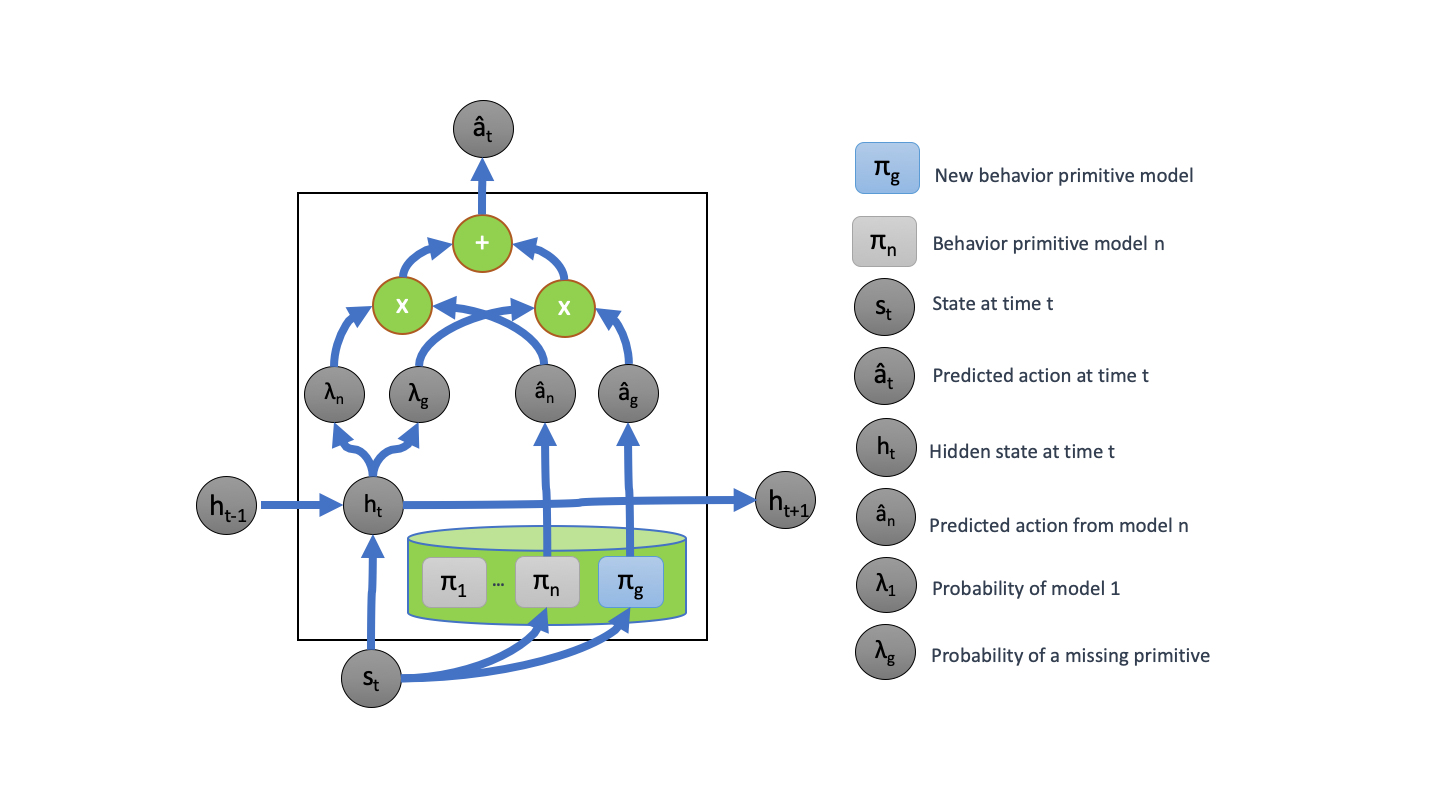}
  \caption{Hierarchical recurrent deep network architecture for task decomposition, novel behavior primitive discovery, and behavior blending.  }
  \label{fig:rnn}
\end{figure}

Figure \ref{fig:rnn} illustrates how we compute the predicted action $\hat{a}_t$ at time $t$.  In the figure, the probability of $\pi$ given state $s_{\rho_t}$ is $\lambda_\pi = p(\pi|s_{\rho_t})$ for $\pi \in \mathcal{B}$.  The latent representation $h_t$ at the current timepoint $t$ is a function of both the value of the latent representation of the previous state $h_{t-1}$ and the current state $s_t$

Figure \ref{fig:model} details the architecture used for \ourmethod based on the Husky+UR5 dataset example.  Unless otherwise specified, the fully connected (FC) layers have ReLU activations, except for the output layers from behavior primitive models.  The last layer of behavior primitive models have linear activations to support diverse action predictions.  While not shown in Figure \ref{fig:model}, the network architecture also returns the predicted latent embedding and behavior primitive distribution for additional visualization and analysis.  

\subsection{Discovering and Training New Behavior Primitives}

An important aspect of our approach is the ability to discover and create new behavior primitives from a set of trajectories and a partial behavior primitive library.  \ourmethod detects and trains new behavior primitive models simultaneously.  As shown in figure \ref{fig:rnn}, \ourmethod supports building new behavior primitive models by adding additional randomly initialized behavior models to the library prior to training.
 For our experiments, we assume that we know the correct number of missing primitives.

We define a \emph{gap} in a trajectory as region within a demonstration where actions are not predicted with high probability using the existing behavior primitive models.
A gap in a trajectory implies that the current library of behavior primitives is insufficient to describe a set of state-action tuples $\rho$ in some part of the given trajectory.  This also implies that the probability $p(\pi|\rho_t)$ that the data $\rho_t$ for time point $t$ was generated by the current library of behavior primitive models is low  for all $\pi \in \mathcal{B}$.  These low probabilities increase the likelihood that an additional randomly initialized behavior primitive policy $\pi_{new}$ might have a higher probability $p(\pi_{new}|\rho_t)>p(\pi|\rho_t)$ for $\pi \in \mathcal{B}$. The data $\rho_t$ is then used to train $\pi_{new}$. For nearby data in the same gap region $\rho_{t+1}$, it is now more likely that $p(\pi_{new}|\rho_{t+1})>p(\pi|\rho_{t+1})$ for $\pi \in \mathcal{B}$.
This mechanism allows $\pi_{new}$ to develop in to a new behavior primitive that is not well covered by existing primitives.

\subsection{Training Details}
\ourmethod is trained end-to-end by back propagation. This is possible because all functions in the model are differentiable with the exception the \texttt{argmax} function.  For experiments making use of pretrained behavior primitive models, the contents of the behavior primitive library are trained using the DART~\cite{DART} technique for imitation learning.

As shown in Equation \ref{mse}, the loss used to train the model is mean squared error between the predicted and observed actions over all timepoints and all demonstrations.  There is no loss term for label prediction accuracy, because we assume that the demonstrations are unlabeled.

\subsection{Metrics}

Two metrics are computed to estimate performance.  First, we evaluate mean squared error (MSE) as shown in Equation \ref{mse} between the predicted and given action. 
Second, we compute behavior primitive label accuracy which is a comparison between the predicted and given behavior primitive label. Label accuracy is computed as the number of matching labels divided by the total number of comparisons. Both metrics are computed over all timepoints and over all demonstrations in the test set.

\subsection{Baseline Implementations}
Shiarli et al. \cite{TACO18} developed TACO, which aligned subtasks to demonstrations given a library of primitives and a \emph{task sketch}, where a task sketch describes the sequence in which subtasks will appear. In addition, in their recent work \cite{TACO18}, they extended the connectionist temporal classification (CTC) algorithm \cite{CTC06}, commonly used to align sequences for speech recognition, for use with identifying subtasks. For this work, we use TACO and the extended version of CTC as baseline comparisons for our algorithm, using an open source implementation~\footnote{https://github.com/KyriacosShiarli/taco}. Both were tested using MLP and RNN architectures.

\section{EXPERIMENTS AND DISCUSSION}
We evaluate \ourmethod using a reach-grab-lift task using a Husky+UR5 environment.  The dataset consists of 100 demonstrations of a Clearpath Husky robot with a UR5 manipulator performing a variety of reach, grasp, and lift tasks, see Figure~\ref{fig:husky}. The number of time steps in the demonstrations varied from 1000 to 1800, but each used all three primitives: reach, grasp, and lift. 

The first experiment quantifies the ability of \ourmethod to identify primitive task labels from demonstration independently from learning behavior primitives.  The second experiment evaluates the ability of \ourmethod to identify parts of demonstrations that are not represented by existing behavior primitives and rebuild the missing behavior primitive.

\subsection{Reconstruction from existing primitives}

Our initial experiment is an ablation study that separately evaluates the estimate of the primitive behavior probability distribution and the action predictions from learning behavior primitives.  We train and freeze behavior primitive models for \emph{reach}, \emph{grasp}, and \emph{lift} using the ground truth labeled data from trajectories.  We evaluated \ourmethod, TACO \cite{TACO18}, and CTC based on label classification accuracy.  For Taco and CTC we additionally compared the methods using MLP and RNN based underlying network models.  We evaluated all methods based on an 80/20 split of demonstrations into training and test sets.  The average of five independent runs were obtained for each approach.  In Table \ref{table:husky}, we show the results of the comparison.

\begin{figure}[ht]
\centering
  \subfloat[Sample trajectory label accuracy]{
  \includegraphics[width=.75\linewidth]{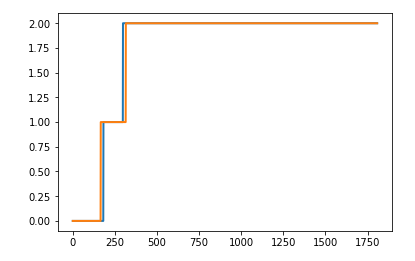}
   }\break
   \subfloat[Missing primitive label accuracy]{
   \includegraphics[width=.75\linewidth]{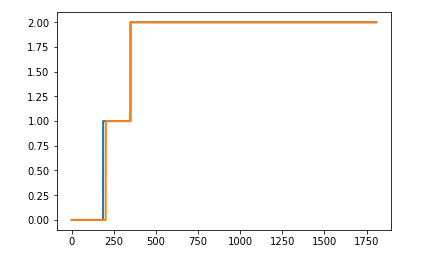}
   }
  \caption{Example behavior primitive label accuracy for a single test demonstration.  We compared the label predictions given by \ourmethod (red) to the ground truth (blue) (a) A sample reconstruction for a single trajectory with an existing behavior primitive library.  Timepoints are on the x-axis. and behavior primitive label is on they y-axis.  The labels 0,1, and 2 correspond to reach, grasp, and lift respectively. (b) Reconstruction of an example trajectory and discovery of a missing behavior primitive (grasp).  }
  \label{fig:reconstruction}
\end{figure}

Figure \ref{fig:reconstruction}(a), shows a comparisons between between the predicted label based on Equation \ref{eqn:label} and the ground truth label.  Over all trajectories in the test set, the average label classification accuracy was 96\% compared to the ground truth label.  The summary of results are shown in Table \ref{table:husky}.

\subsection{Behavior Primitive Discovery}

\begin{figure}[ht]
\centering
  \subfloat[behavior label accuracy]{
  \includegraphics[width=.75\linewidth]{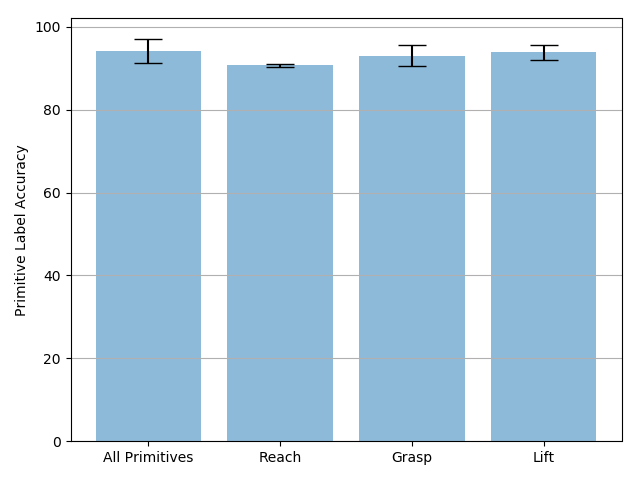}
   }
   \break
   \subfloat[action prediction MSE]{
   \includegraphics[width=.75\linewidth]{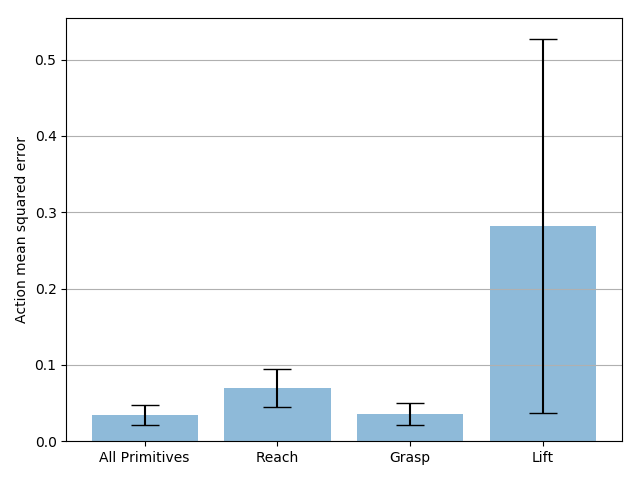}
   }
  \caption{Accuracy of \ourmethod to correctly identify a primitive's label on the validation set (twenty randomly selected trajectories). (a) The leftmost bar shows performance when all primitives are in the library, successive bars denote accuracy when the \emph{reach}, \emph{grasp}, and \emph{lift} primitives are dropped out and learned from a randomly generated ``gap" primitive. Error bars represent the standard deviation across five validation trials. (b) Mean squared error between the ground truth action and the learned model's estimate averaged across twenty randomly selected test trajectories five times.  }
  \label{fig:labelling}
\end{figure}

In our next experiment, we evaluate the ability of \ourmethod to recognize and build a missing behavior primitive model.  We ran a leave-one-behavior-out experiment where one of the three primitives (i.e. reach, grasp, lift) was replaced with a randomly-initialized behavior primitive.  This experiment used the same 100 trajectories on the Husky+UR5 dataset discussed in the previous section and a 80/20 split between training and validation sets. Again, five trials were run with the training and validation sets randomly chosen. The label accuracy and action prediction MSE are shown in \ref{fig:labelling}. The leftmost bar shows the results with all primitives pre-trained with behavior cloning. The remaining bars show the accuracy when reach, grasp and lift, respectively, were replaced with the gap primitive. Note, the gap primitive was updated throughout the training with back-propagation such that the final primitive ideally would perform as well as the original pre-trained, behavior-cloned version; this comparison is shown with the action prediction MSE. The error bars show the standard deviation across the five trials. While the label accuracy across all three replaced primitives is approximately the same, the action prediction for the lift primitive is significantly worse. We believe this is due to the larger variance in lift trajectories. Unlike the reach and grasp which have restrictions placed on their final target position (it needs to be near the block), the final position of lift is randomly placed above the block's starting position.

As shown in the sample trajectory in Figure \ref{fig:reconstruction}(b), the label prediction of the trained model closely aligns with the ground truth label from the example trajectory.  Over all of the test trajectories, the average label classification accuracy was 96\%.

\subsection{Visualizing the Learned Latent Space}

To better understand the role of the embedding space for predicting the primitive probability distribution, we visualized the embedding of all states vectors from the test set in the recurrent hidden layer.  We would expect that a useful latent embedding would naturally cluster states that correspond to different primitives into distinct locations in the embedding space.
\begin{figure}[ht]
\centering
  \includegraphics[width=.75\linewidth]{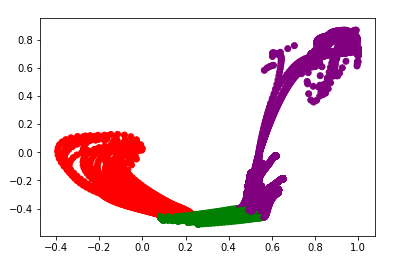}
  \caption{The organization of the learned latent space associated with the Husky-UR5 dataset for reach, grasp, and lift (red, green, and purple respectively).    }
  \label{fig:latent}
\end{figure}
Figure \ref{fig:latent} shows layout of the latent space in two dimensions.  Each point corresponds to a state vector from the test dataset.  The points are colored by the ground truth label.

\subsection{Jaco Dial Domain Dataset}
\begin{figure}[ht]
\centering
  \includegraphics[width=.75\linewidth]{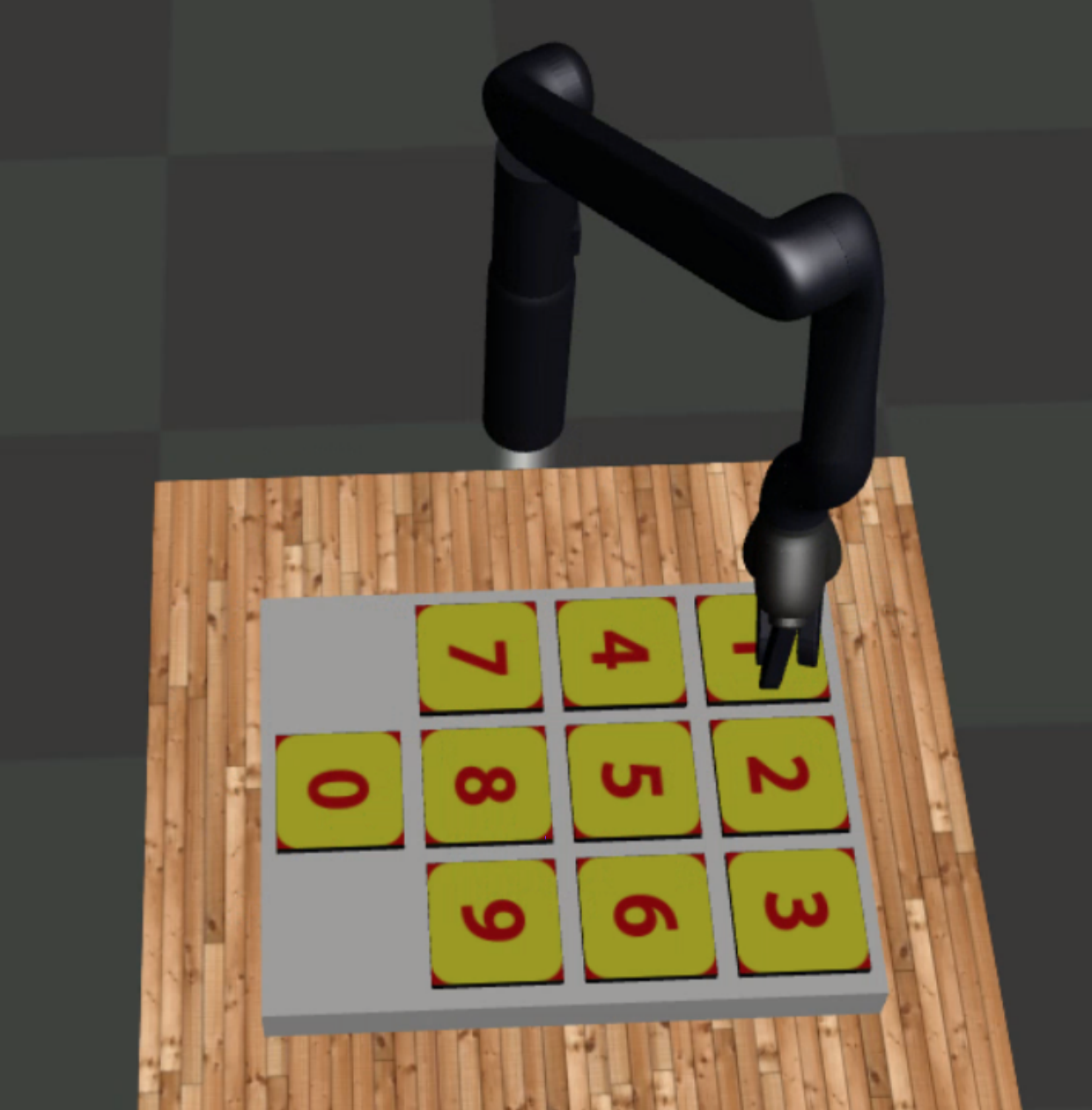}
  \caption{The joint-domain dial scenario. A Jaco manipulator modeled in Mujoco presses a sequences of 4 keys on a dialpad.  The positions of the keys are randomly shuffled for each demonstration.  The positions of the joints and positions of the keys are given as state information. }
  \label{fig:pinpad}
\end{figure}

We also make use of the Jaco dial domain dataset\cite{TACO18} illustrated in Figure \ref{fig:pinpad}.  The dial dataset is composed of demonstrations from a Jaco manipulator pressing 4 keys in sequence (e.g. 3,5,4,7).  The positions of the keys are randomly shuffled for each demonstration, but the position of each key is given in the state vector.  The intention is to treat pressing an individual digit as a behavior primitive. For this dataset, label prediction accuracy is a challenging metric without a task sketch because the starting position of the jaco may not provide clues about which button will be pressed.  As the jaco gets closer to a button, it becomes more clear which button will be pressed.  The dataset of dialpad demonstrations were generated using default parameters and code from TACO\cite{TACO18}.

\subsection{Dial Domain Comparison}
The goal of this comparison is to evaluate the label prediction accuracy of the metacontroller in \ourmethod.  To isolate the label predictions of the metacontroller, the behavior primitive library is pretrained on the training dataset  including 1200 demonstrations and frozen.  Label classification and action prediction accuracy is then evaluated on the test set including 280 demonstrations.

The average results of 5 runs are shown for TACO and CTC.  We evaluate each approach using the same label accuracy and action prediction metrics.  The summary of results are shown in Table \ref{table:pinpad}.  We found that our approach achieves the highest label accuracy at 65\%.  The overall label accuracy of \ourmethod on the dial dataset is lower than the Husky+UR5 dataset.  Additional analysis revealed that many of the mislabeling occurred at the beginning of a new key press where context about where the Jaco is moving next is weakest.  The dataset is also more challenging than than the Husky dataset because the number of unique behavior primitives has increased from 3 to 10.  

Also of note, we compare our results to TACO which is a weakly supervised approach.  TACO is given the ordering of tasks.  For task sequences of length 4, this means that a random baseline would be expected to achieve an accuracy of 25\%.  For an unlabeled approach like \ourmethod, any of the 10 behavior primitives could be selected at each timepoint.  This means that unlabeled demonstrations the expected accuracy of a random baseline would be 10\%.

\section{CONCLUSION}

In this paper, we describe \ourmethod, an approach to learn behavior primitives from unlabeled demonstrations and a partial set of behavior primitives.  We optimize a metric that directly minimizes reconstruction error for a set of demonstrations using sequences of behavior primitives. We directly compare our results to similar approaches using demonstrations generated from simulations of two different robotic platforms and achieve both better label accuracy and reconstruction accuracy as measured by action prediction mean squared error. While we have demonstrated success in these tasks, there are limitations to our approach.  The number additional primitives to add to the library must be decided prior to training. In spite of these limitations, we believe that \ourmethod is a useful contribution to the community that may be relevant in a number of different domains.

\begin{table}[]
\caption{Method comparisons using the Husky UR5 Reach and Grasp dataset.}
\label{table:husky}
\begin{tabular}{|c|c|c|}
\hline
Husky UR5  & Label Accuracy & MSE Action Prediction  \\ \hline
\ourmethod    & \textbf{96\%}    &  \textbf{0.053} \\ \hline
TACO (MLP)   & 74\% &  3.59 \\ \hline
TACO (RNN)   & 73\% &  3.75 \\ \hline
CTC (MLP) & 25\% & 4.20 \\ \hline
CTC (RNN) & 33\% & 2.68 \\ \hline
\end{tabular}
\end{table}

\begin{table}[]
\caption{Method comparisons using the Jaco Pinpad dataset. *TACO (RNN) resulted in NaN loss after repeated attempts.}
\label{table:pinpad}
\begin{tabular}{|c|c|c|}
\hline
Jaco Pinpad & Label Accuracy & MSE Action Prediction  \\ \hline
\ourmethod  & \textbf{65\%}    &  \textbf{0.0061} \\ \hline
TACO (MLP)   & 47\% &  0.55 \\ \hline
TACO (RNN)   & * &  * \\ \hline
CTC (MLP) & 31\% & 0.57 \\ \hline
CTC (RNN) & 29\% & 0.58 \\ \hline
\end{tabular}
\end{table}



\bibliographystyle{unsrt}
\bibliography{references}

\end{document}